# Exploiting the Power of Human-Robot Collaboration: Coupling and Scale Effects in Bricklaying*

Jia-Rui Lin, and Minghui Wu

*Abstract*— **As an important contributor to GDP growth, the construction industry is suffering from labor shortage due to population ageing, COVID-19 pandemic, and harsh environments. Considering the complexity and dynamics of construction environment, it is still challenging to develop fully automated robots. For a long time in the future, workers and robots will coexist and collaborate with each other to build or maintain a facility efficiently. As an emerging field, human-robot collaboration (HRC) still faces various open problems. To this end, this pioneer research introduces an agent-based modeling approach to investigate the coupling effect and scale effect of HRC in the bricklaying process. With multiple experiments based on simulation, the dynamic and complex nature of HRC is illustrated in two folds: 1) agents in HRC are interdependent due to human factors of workers, features of robots, and their collaboration behaviors; 2) different parameters of HRC are correlated and have significant impacts on construction productivity (CP). Accidentally and interestingly, it is discovered that HRC has a scale effect on CP, which means increasing the number of collaborated human-robot teams will lead to higher CP even if the human-robot ratio keeps unchanged. Overall, it is argued that more investigations in HRC are needed for efficient construction, occupational safety, etc.; and this research can be taken as a stepstone for developing and evaluating new robots, optimizing HRC processes, and even training future industrial workers in the construction industry.**

## I. Introduction

As a pillar industry of the national economy, the construction sector builds and maintains buildings and infrastructures for the support of production, living, transportation, and healthcare. Despite this, the construction industry has always been criticized for its harsh working environment, occupational accidents, and low productivity. The willingness of workers to stay in the construction industry continues to be low, and the labor shortage is increasing. Meanwhile, the trend of global population ageing and the COVID-19 pandemic further impact the fragile labor market supply in the construction industry. Therefore, development of new construction methods, i.e., construction robots, modular construction, have become the cutting-edge trends in the construction industry[1].

However, the construction process is highly complex and dynamic. And, the construction site is a typical unstructured scene, where the surrounding, layout and the appearance of building components change frequently. Therefore, compared with the development and deployment of robots in structured scenarios of manufacturing, developing fully automated construction robots in an unstructured and dynamic construction site is still challenging[2]. It is foreseeable that for a long time, the construction site will be in a state where humans and robots coexist and collaborate with each other.

Generally, workers and robots in the construction process have their own characteristics and advantages. For example, workers are highly adaptable and have a strong ability to cope with complex environments, while they have specific limitations in terms of ergonomics and occupational safety. They are limited by human factors such as forgetting and fatigue, as well as risk constraints such as human-machine collision and harsh environments, which leading to unstable performance in quality and efficiency. On the contrary, robots have relatively stable performance in terms of quality and efficiency, and is applicable in extreme environments. However, their capacity is limited in complex and dynamic scenarios due to insufficient adaptability. How to effectively utilize the advantages of workers and robots to achieve effective collaboration between the two and maximize construction efficiency and quality is a key challenge for human-robot collaboration in the construction industry. To overcome this challenge, a series of open problems should be addressed through tight collaboration with experts from area of construction, robotics, informatics, and ergonomics, etc.

Currently, research on construction robots still focuses on the development of robots and algorithms for specific construction tasks, which aim to improve the performance of a single robot[3]. Less attention is paid to the worker side, and there still lacks investigations on HRC at a macro level.

To this end, this research conducts a pioneer exploratory work in HRC for the construction industry. Due to the complexity, high-risk and irreproducibility of the real site experiments, this study introduces an agent-based modeling and simulation approach, to systematically analyze the interdependency of human factors, robot features, and HRC parameters as well as their impacts on construction productivity (CP) in the bricklaying scenario.

## II. Methodology

In view of the flexibility of the agent-based modeling and simulation[4], this study introduces an agent-based HRC modeling and simulation approach (Fig. 1), which consists of four steps: parameter extraction, agent-based modeling,

*Research supported by National Natural Science Foundation of China (No. 51908323, No. 72091512) and the National Key R&D Program of China (No. 2018YFD1100900).

Jia-Rui Lin is with the Department of Civil Engineering, Tsinghua University, Beijing 100084, China (corresponding author, phone: 0086-6278-9225; e-mail: lin611@tsinghua.edu.cn).

Minghui Wu was with the Department of Civil Engineering, Tsinghua University, Beijing 100084, China. He is now with the Department of Civil and Environmental Engineering, University of Michigan, Ann Arbor, MI 48109, USA (e-mail: minghuiw@umich.edu).

experimental design, and analysis and insights. The following will explain each step in detail.

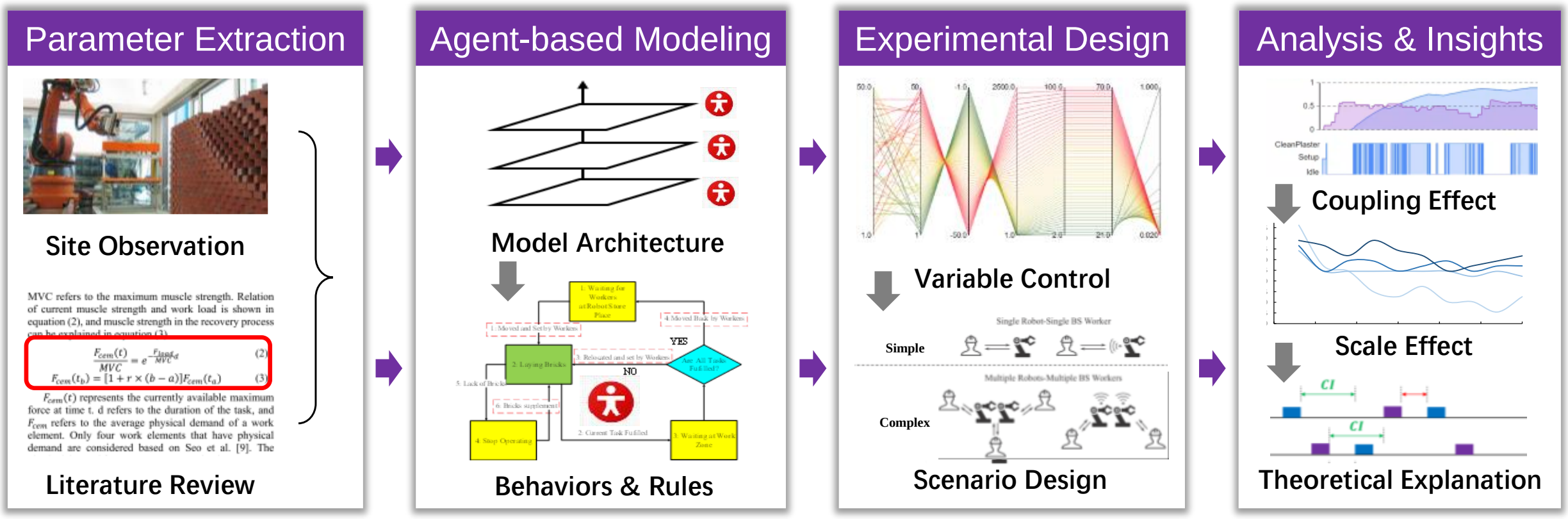

Figure 1. Proposed methodology

## A. *Parameter Extraction*

This step deals with the acquisition of simulation parameters, and is mainly achieved from two aspects. On one hand, observation of real site experiment is conducted, video capture or other techniques are utilized to record the process of HRC. Later, humans and robots are collaborated can be extracted from the recorded data. These may include the sequence of tasks of workers and robots, performance of robots, frequency and duration of human-robot interaction, etc. on the other hand, literature review is also adopted to extract data from published articles. For example, CP of robot-based bricklaying, and ergonomic models such as forgetting and fatigue are obtained from literatures [5] and [6].

## B. *Agent-based Modeling*

Based on the parameters extracted above, Anylogic, a widely used modeling software, is adopted to create the agent-based model for HRC. The developed model mainly includes four kinds of agents: workers, robots, bricks, and site environment recorders. The main relationships of them are illustrated in Fig. 2. In general, the following parameters are considered when modeling the HRC process in bricklaying.

- Environment: location and capacity of temporary storage, location and length of walls, etc.;
- Worker: role, position, speed, mortar removing performance, fatigue and forgetting model, etc.;
- Robot: position, movement speed, bricklaying performance, brick capacity, safety space, etc.;
- Collaboration: check interval (CI) that for workers to check the robot, and supply limit (SL) when the worker starts supplying bricks, proactive interaction, etc.

More details could on parameter definition and HRC modeling can be found in our papers [7] and [8].

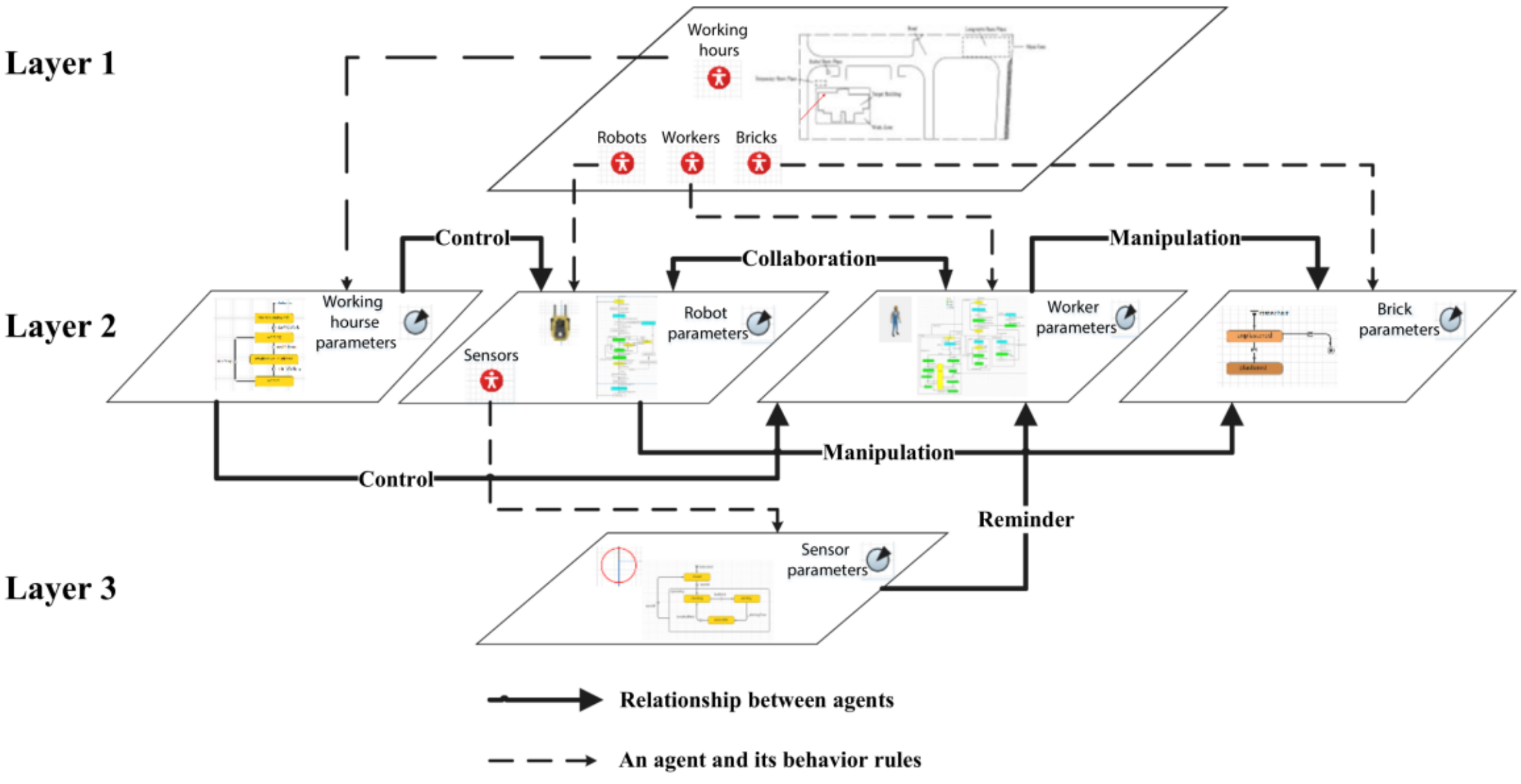

Figure 2. Developed agent-based model for human-robot collaboration

### C. *Experimental Design*

In this step, a series of simulation experiments are designed based on control variable method. That is, simulations based on different CIs, SLs, and different human-robot interaction modes, etc., are created and implemented to generate data for future analysis. Meanwhile, different HRC scenarios, i.e., single robot-single worker scenario, multiple robot-multiple worker scenario, are also designed and simulated.

### D. *Analysis and Insights*

At last, generated data from simulations are visualized, analyzed to obtained insights on how human and robots are collaborated. Meanwhile, theoretical model is also developed to explain the underlying mechanism of HRC.

## III. RESULTS

Based on the proposed method, many simulation experiments were carried out in this study. By data analysis and theoretical explanation, the following results related to the interdependency of HRC parameters and the impacts of HRC on CP were obtained. Note that CP is estimated based on the construction time of a certain quantity of work.

### A. *Dynamics and Coupling Effect of HRC*

According to Fig. 3, the behavior and workload of different workers are significantly different due their differences in roles. The worker who removes the mortar (EMR worker) is always in a state with intensive workload because his/her performance is lower than the performance of the robot for bricklaying; while the brick supplement worker (BS workers) often need to repeatedly check whether the bricks should be supplied. If the check interval is not selected properly, there will be many redundant checking tasks. In addition, due to the safety distance defined between the robot and the worker, the low efficiency of the EMR worker may hinders the bricklaying process of the robot, resulting in intermittent interruptions time by time.

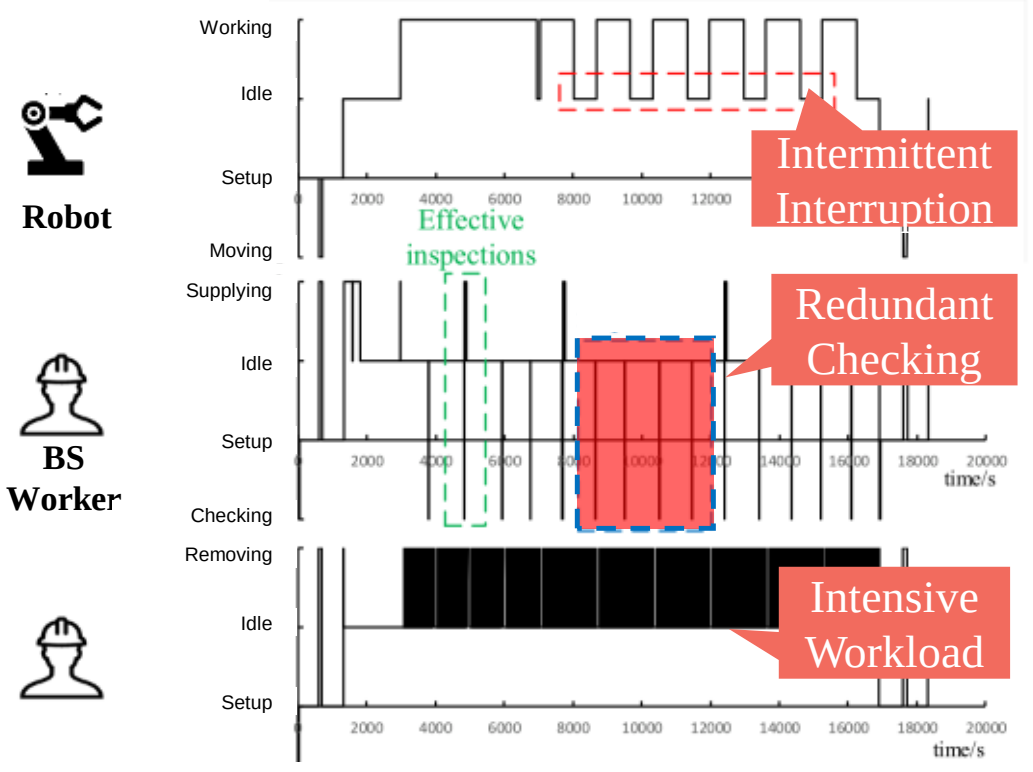

Figure 3. State changes of workers and robot

The experiment also shows that when the EMR worker enters a state of fatigue after a long time of intensive work, his/her efficiency will drop significantly, which may also lead to intermittent interruptions of robots (Fig. 4). Note that to demonstrate the results in Fig.3 and Fig. 4, workers are set to work for a long time, i.e., more than 8 hours a day.

In summary, states of workers and robot are interdependent, and the performance of workers or robot are influenced by different parameters, making it difficult to understand and optimize the HRC process and improve the productivity.

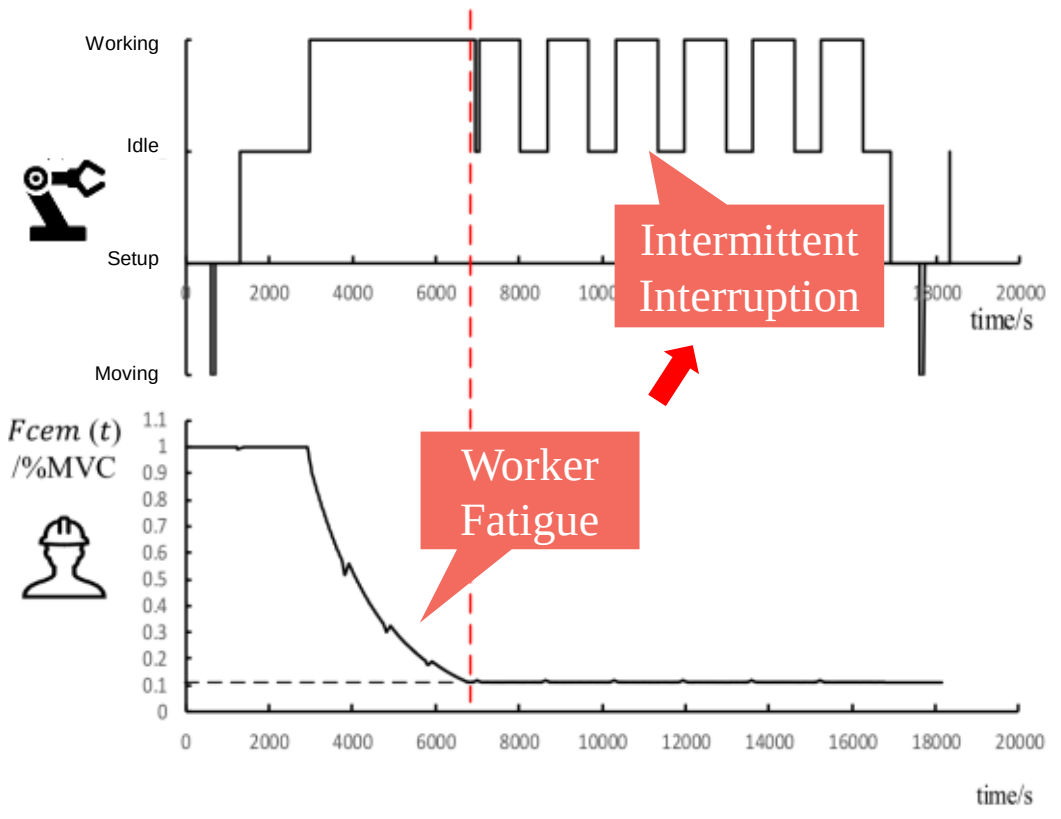

Figure 4. Impact of worker fatigue on the performance of robot

### B. *Impact of HRC on Construction Productivity*

Based on the simulation data of different scenarios, this research systematically analyzed the effect of different HRC parameters (including SL, CI, interaction modes) on CP.

For the single robot-single worker (SRSW) scenarios, when SL are in different ranges, the relationship between CI and CP follows different rules. As shown in Fig. 5(a) and Fig. 5(b), when the value of SL is large, the total construction time increases linearly as CI increases. However, when the value of SL is small, although the total construction time increases as CI increases, there is a flattening or oscillation range in the middle. Meanwhile, for the multiple robot-single worker (MRSW) scenarios in Fig. 5(c) and Fig. 5(d), the relationship between CI, SL and CP are quite different comparing to SRSW scenarios. That is, when the SL is small, the construction time decreases as CI increases; while when the SL is large, the construction time increases as CI increases.

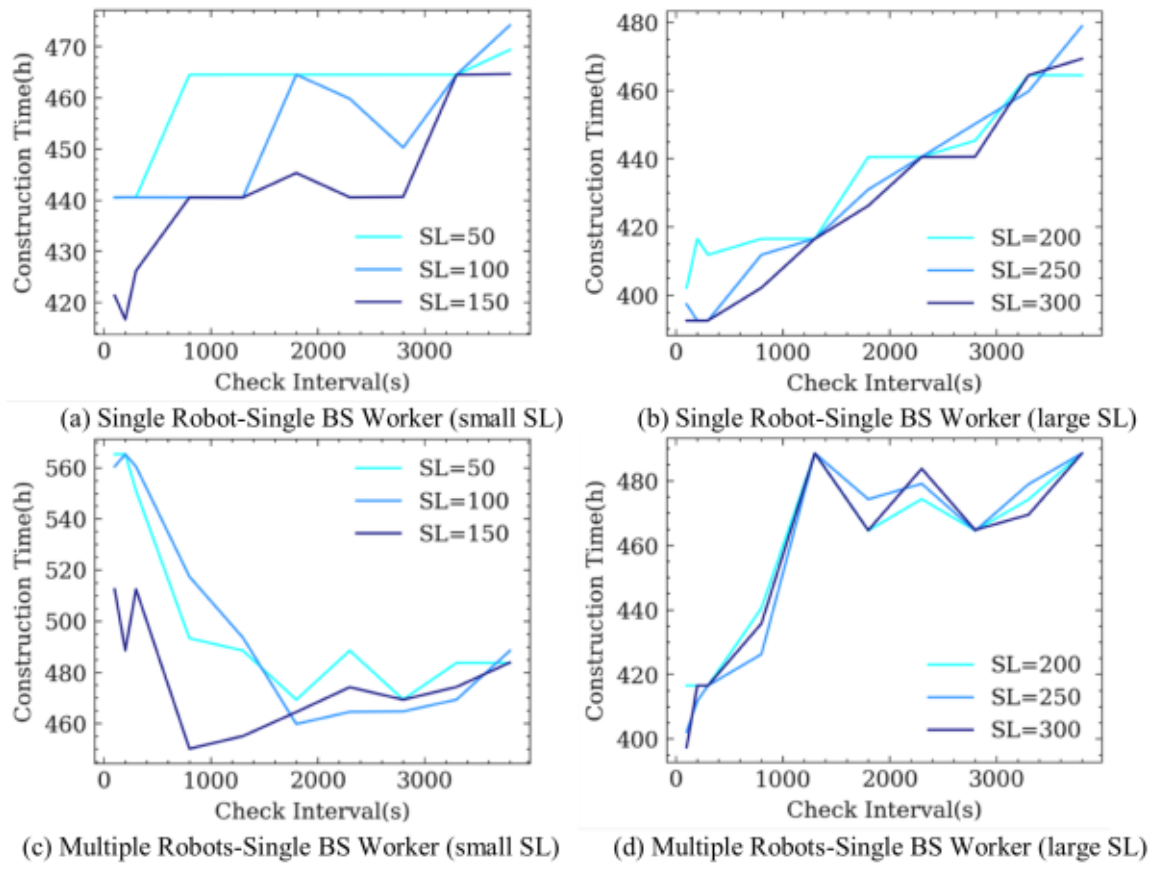

Figure 5. Impacts of CI and SL on productivity

The analysis shows that there are complex interdependencies and constraints between various parameters of the HRC process, which may have different effects on CP. This also implies that parameters of HRC are tightly coupled with each other, and more in-depth investigations are needed to clearly understand and further optimize the HRC processes.

Except for the scenarios where the workers passively check the status of robots, we also introduced sensors to help robot proactive feedback its states to the workers. As show in Fig. 6, comparing to passive interaction mode, proactive interaction mode between workers and robot can significantly improve the CP of both SRSW scenarios and MRSW scenarios. In most cases, the CP can be improved by more than 20%. This shows that the active communication and interaction between robots and workers has a very positive significance for improving efficiency, which means that robots with self-awareness are much better than traditional ones.

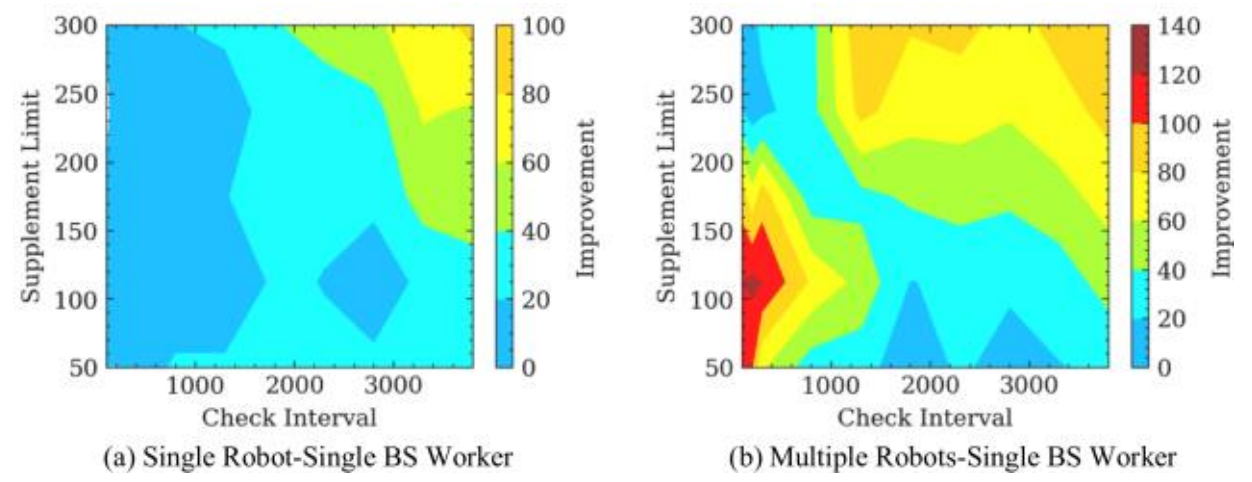

Figure 6. Productivity improvement through proactive interaction

### C. *Scale Effect of HRC*

At last, scenarios related to multiple robot-multiple worker (MRMW) are also considered and investigated. According to Fig. 7, the general distribution of construction time in MRMW scenario is similar to the SRSW scenario. Though dark blue area is similar between the, some green area in Fig. 7(a) is replaced by the blue area in Fig. 7(b). This indicates that productivity increases although the ratio between robots and workers remains the same. That is, there exists a scale effect between the collaboration of robots and BS workers, especially for scenarios with larger CI. This implies that largescale promotion and application of construction robots on construction sites may increase the performance of construction dramatically.

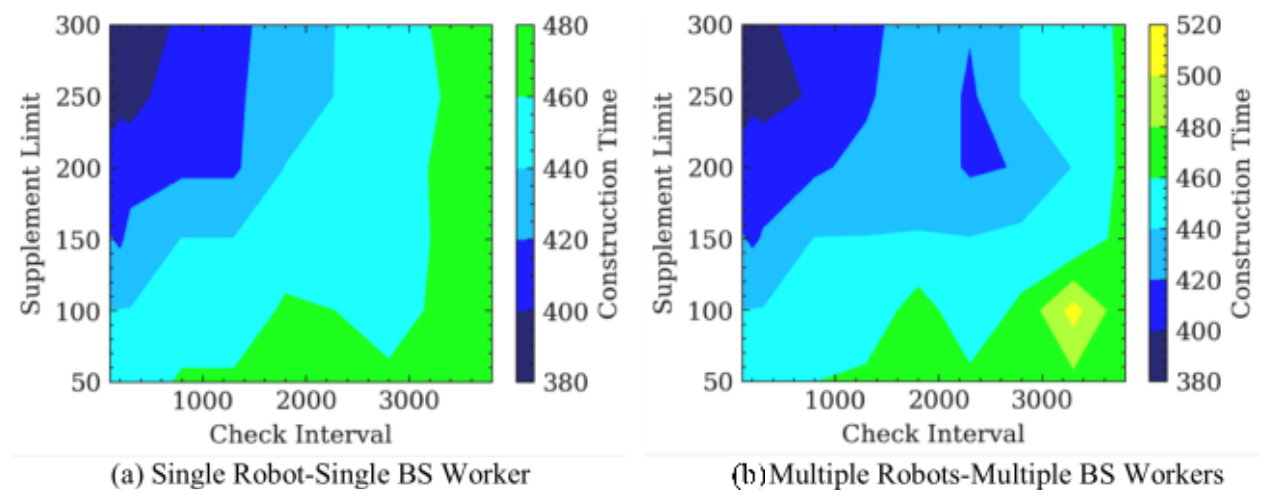

Figure 7. Scale effect when increasing the number of human-robot teams

The underlying mechanism behind scale effect is that the large portion of overlap between check intervals makes that the general checking interval (GCI) smaller than the CI determined by the supplement strategy (Fig. 8). In other words, when there are two human-robot teams working together, though the CI of each BS worker keeps unchanged, they can make GCI smaller by helping their partners check the robot. Thus, each robot can be checked in a shorter interval, or GCI in this case. Here we defined this mechanism as mutual help of workers, which could explain the scale effect theoretically.

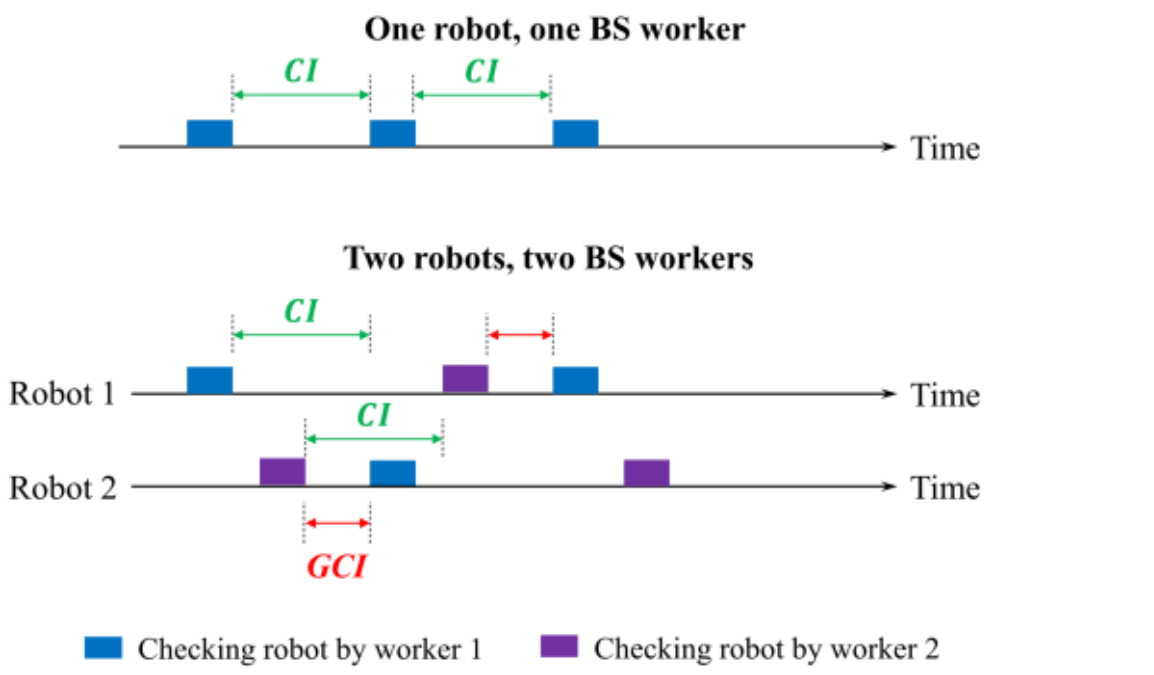

Figure 8. Mutual help of workers – a theoretical explanation of scale effect

## IV. Conclusions

In this research, an agent-based modeling and simulation approach is introduced to investigate the human-robot collaboration (HRC) process and its impact on construction performance. With the developed model and various experiments, it is showed that: 1) multiple parameters related to HRC, from the human side, the robot side and their interactions, are highly interdependent and coupled with each other; 2) impacts of HRC on construction productivity are complex, parameters should be carefully chosen and proactive interaction mode is recommended; 3) HRC has a scale effect on productivity due to mutual help mechanism between workers, and large scale adoption robots may get a much better performance than linearly adding multiple human-robot teams.

Overall, this research contributes an integrated approach to simulate the HRC process and evaluate its impacts on construction productivity. we also demonstrate a new paradigm to understand the HRC process in various robot-based construction scenarios, which can be taken as a stepstone for the evaluation and development of new robots, optimization of HRC process to maximize construction performance and occupational health, and even training of skilled workers for future construction.